\documentclass{article}

\usepackage{graphicx}
\usepackage[numbers]{natbib}
\usepackage{makecell}
\usepackage{hyperref}
\usepackage{amsmath,amssymb,amsthm,enumerate,mathrsfs}

\theoremstyle{definition}

\numberwithin{equation}{section}

\begin{document}
	
	\makeatletter

	\begin{center}
		\large{\bf Evaluating the Performance of ChatGPT for Spam Email Detection}
	\end{center}\vspace{5mm}
	
	\begin{center}
		\textsc{Shijing Si, Yuwei Wu\footnote{Co-first authors: Shijing Si and Yuwei Wu. This work was supported
				by National Natural Science Foundation of China (72071098),  the Fundamental Research Funds for the Central Universities under Grant (2022114012, 41004885),
				and the High-End Foreign Expert Recruitment Program of the Ministry of Science and Technology of the People's Republic of China, the construction of cutting-edge courses and academic research plans of International Finance and Big Data Management, under Grant G2023138001.}, {Le Tang}\footnote{Corresponding author: Le Tang, 1067061406@qq.com.}, Yugui Zhang, Jedrek Wosik and Qinliang Su}
	\end{center}
	

	\vspace{2mm}

	\footnotesize{
		\noindent\begin{minipage}{14cm}
			{\bf Abstract:}
			Email continues to be a pivotal and extensively utilized communication medium within professional and commercial domains. Nonetheless, the prevalence of spam emails poses a significant challenge for users, disrupting their daily routines and diminishing productivity.
			Consequently, accurately identifying and filtering spam based on content has become crucial for cybersecurity. Recent advancements in natural language processing, particularly with large language models like ChatGPT, have shown remarkable performance in tasks such as question answering and text generation. However, its potential in spam identification remains underexplored.
			To fill in the gap, this study attempts to evaluate ChatGPT's capabilities for spam identification in both English and Chinese email datasets.
			We employ ChatGPT for spam email detection using in-context learning, which requires a prompt instruction with (or without) a few demonstrations. We also investigate how the number of demonstrations in the prompt affects the performance of ChatGPT. For comparison,
			we also implement five popular benchmark methods, including naive Bayes, support vector machines (SVM), logistic regression (LR), feedforward dense neural networks (DNN), and BERT classifiers. Through extensive experiments, the performance of ChatGPT is significantly worse than deep supervised learning methods in the large English dataset, while it presents superior performance on the low-resourced Chinese dataset. This study provides insights into the potential and limitations of ChatGPT for spam identification, highlighting its potential as a viable solution for resource-constrained language domains.
		\end{minipage}
		\\[5mm]
		
		\noindent{\bf Keywords:} {ChatGPT, Large Language Models, Spam email detection, Cybersecurity, In-context Learning}

		\hbox to14cm{\hrulefill}\par


		\section{Introduction}
		
		Used as a swift and cost-efficient communication tool, email underpins global connectivity and is fundamental in various professional and personal activities, such as international electronic commerce \citep{hudak2017importance}. As of 2020, the international email user base was approximately 4.037 billion. Due to its very nature of accessibility, email inevitably compromises on privacy \citep{ackerman1999privacy}, hence, exploited by corporations, organizations, and criminals alike for marketing and fraudulent purposes.
		
		Spam content predominantly encompasses deceptive information or advertisements \citep{frans2023review}, and occasionally includes harmful elements such as computer viruses, explicit material, or adverse information. The rampant spread of spam not only poses a nuisance to individuals but also instigates a squandering of network resources, in addition to eliciting a myriad of security threats \citep{AhmedAAKAS22}. As such, the development of an efficacious model capable of accurately identifying and consequently filtering spam is of paramount importance \citep{SamarthraoR22enhance}.

		Numerous methods have been established in the realm of spam detection and classification \citep{KarimASKA19,Akinyelu21}. Conventional approaches encompass techniques such as character matching, the naive Bayes algorithm \citep{kumar2020predictive}, support vector machines (SVM) \citep{sharma2016mail} and ensemble learning \citep{BhardwajS23}. In the wake of deep learning advancements, a variety of deep models, including Convolutional Neural Networks (CNN), Recurrent Neural Networks (RNN), and BERT, have been employed in spam recognition \citep{seth2017multimodal,fazzolari2021experience,ManasaMB24}. For instance, \cite{DoshiPSS23} proposed a comprehensive dual-layer architecture for phishing and spam email detection. \cite{ZavrakY23} developed a hierarchical attention hybrid deep learning method for spam email detection.
		
		Recently large language models (LLMs) such as ChatGPT have shown impressive performance in many natural language processing tasks \cite{radford2019language}, including question answering, text generation, etc. In-context learning \cite{garg2022what,wei2022cot}, central to these large models, refers to their prediction capabilities given a specific context; these models analyze and generate words or sentences based on the instruction (i.e., prompt), which aids their ability to coherently operate in tasks such as conversation, text completion, translation, and summarization. However, while ChatGPT has been used in various tasks of text processing, little research has been done on spam detection. To fill in the gap,
		we aim to explore how well ChatGPT performs at spam detection via in-context learning. We design experiments to
		systematically evaluate the performance of ChatGPT in both English and Chinese email datasets.

		Our contributions can be summarized as follows:
		\begin{itemize}
			\item We compile a low-resourced Chinese spam email datasets, which can be used in the community.
			\item We evaluate the performance of ChatGPT on spam email detection via in-context learning. From our experiments, the performance of ChatGPT is worse than deep supervised models in spam detection, but it has shown benefits in the low-resourced setting (Chinese dataset).
			\item We also investigate how the number of instances affects the performance of in-context learning.
		\end{itemize}
		
		The remainder of the paper is organized as follows. Section 2 covers the related works of this research. Section 3 presents the experimental setup in detail. Section 4 exhibits the experimental results and analyses. Section 5 is the conclusion.
		
		\section{Related Works}\label{sec:rel}
		\subsection{Spam Classification}
		Based on the types of data used, existing spam classification algorithms can be broadly categorized into two groups:
		The first category involves recognition based on mail features. This approach involves extracting features from mail titles and employing supervised classifiers like the SVM model for classification \citep{khamis2020header}. Some studies have also considered the cumulative relative frequency of each spam feature to enhance the performance of supervised classifiers \citep{fazzolari2021experience}. The feature-based spam recognition dataset is multidimensional, making it suitable for multiple classification models. Consequently, researchers have compared the performance of different machine learning models \citep{awad2011machine}, such as SVM, deep neural networks (DNN), and k-nearest neighbor (KNN) \citep{yeruva2022mail}, in spam recognition tasks.
		
		The second category focuses on email content identification. For instance, the relaxed online SVM model can be utilized to effectively filter and classify mail content \citep{sculley2007relaxed}, naive Bayes classifiers have also demonstrated effectiveness in email classification \citep{ma2020comparative}. In recent years, deep learning algorithms have been increasingly employed in spam recognition. CNNs \citep{seth2017multimodal}, for example, have proven to be effective in identifying spam. The transfer learning of BERT models has also been employed for spam classification, yielding high accuracy  \citep{fazzolari2021experience}. Long Short-Term Memory (LSTM) models have shown better performance on short messages (SMS) datasets compared to machine learning classification algorithms like random forest and SVM \citep{raj2018lstm}.
		
		\subsection{GPT and Its Applications}
		
		GPT, short for generative pre-training, \citep{radford2018improving} is one of the landmark models proposed on the basis of Transformer proposed by Vaswani in 2017 \citep{vaswani2017attention}, another representative model is the BERT model proposed by \citep{devlin2018bert}. On this basis, GPT2 \citep{radford2019language} and GPT3 \citep{brown2020language} have been proposed and widely used. Excellent performance was achieved tasks such as text generation, question and answer, translation, summary, and dialogue.
		
		ChatGPT \citep{ouyang2022training} is an upgraded version of OpenAI in 2022 based on the GPT-3.5 model, which is optimized for conversation tasks, adding input and output of conversation history, and control of conversation policies. ChatGPT excels at conversational tasks and can have natural and fluid conversations with humans.
		
		Spam recognition is a text classification task. ChatGPT generally performs well on text classification tasks, and researchers have demonstrated that the model performs well on some text classification tasks \citep{liu2023summary}. GPT can be used to assist classification decision, by using chatgpt to extract structured and refined knowledge from knowledge graph as the characteristics of classification tasks. \citep{shi2023chatgraph}. It can also be used directly for text classification with appropriate Prompt design, such as for agricultural \citep{zhao2023chatagri}, business news \cite{lopez2023can} and even medical dialogues \citep{wang2023chatgpt}.

		A few research have explored the application of ChatGPT in text binary classification \citep{wang2023chatgpt}, which can provide some reference for spam recognition. In recent studies, ChatGPT was able to identify opinions, sentiment, and emotions contained in the text and categorize emotions. In addition, GPT also performed well in sentiment analysis of disease-related papers \citep{susnjak2023applying} and Twitter texts \citep{sharma2023mining}, indicating that ChatGPT is capable of a wide range of sentiment analysis tasks for texts in different fields.
		
		Due to some uncertainty in the answers of large language models, \citep{borji2023categorical}, when using ChatGPT for text classification tasks, it is necessary to carefully design hints and related strategies to make the answers as stable and effective as possible. If necessary, a small number of samples can be added to guide the model \citep{reiss2023testing}, such as one-shot prompting \citep{korini2023column}, few-shot prompting \citep{zhao2021calibrate}, and chain-of-thought (CoT) prompting design \citep{liu2023gpteval}.
		
		\section{Spam Detection as an Optimization Problem}\label{sec:optimiz}
		
		In this section, we formulate the spam email detection as a binary classification.
		Suppose the training dataset $\{(\textbf{x}_1, y_1), (\textbf{x}_2, y_2), \ldots, (\textbf{x}_n, y_n)\}$ consists of $n$ training examples, each of which has an email $(\textbf{x}_i)$ and label $y_i\in\{0, 1\}$.
		For a text encoder (feature extractor)
		$f_{\mathbf{\theta}}(\textbf{x})$, it extracts hidden representations from the input email $\textbf{x}$ and it can be term frequency--inverse document frequency, BERT encoder \citep{devlin2018bert}, feedforward neural networks \citep{ManasaMB24}, and so on.
		The probability of the $i$-th email being spam is
		\begin{equation}\label{eq:prob.spam}
			P(Y=1|\textbf{x}_i) = \sigma(\textbf{w}^{T}f_{\mathbf{\theta}}(\textbf{x}_i) + b) = \frac{\exp{(\textbf{w}^{T}f_{\mathbf{\theta}}(\textbf{x}_i) + b)}}{1 + \exp{(\textbf{w}^{T}f_{\mathbf{\theta}}(\textbf{x}_i) + b)}},
		\end{equation}
		so we attempt to optimize the cross-entropy loss during the training stage:
		\begin{equation}\label{eq:optim}
			({\mathbf{\theta}}^{*}, {\textbf{w}}^{*}, {b}^{*}) = \arg\min_{\mathbf{\theta}, \textbf{w}, b}\sum_{i=1}^{n}(y_{i}\log\sigma(\textbf{w}^{T}f_{\mathbf{\theta}}(\textbf{x}_i) + b) + (1 - y_i)(1-\sigma(\textbf{w}^{T}f_{\mathbf{\theta}}(\textbf{x}_i) + b))),
		\end{equation}
		where $\hat{\mathbf{\theta}}, \hat{\textbf{w}}, \hat{b}$ are the estimates of $\mathbf{\theta}, \textbf{w}$ and $b$, respectively.
		
		For the large language models (LLMs) like ChatGPT that has more than 100 Billion parameters, we do not need to tune the parameters, but to prompt them directly. The performance of
		a LLM depends on the design of prompts.
		We can tune the prompt on the validation dataset $S_{v}$,
		\begin{equation}\label{eq:prompt.tune}
			\textbf{p}^{*} = \arg\min_{\textbf{p}}\sum_{i\in{S}_{v}}I(\text{LLM}([\textbf{p},~\textbf{x}_i]) \neq y_i),
		\end{equation}
		where $I(\cdot)$ is the indicator function, and it equals to 1 if the output of LLM is not $y_i$, otherwise it equals to 0. The optimization can be over the instruction of prompt or the training examples in the prompt.
		
		\section{Experimental Setup}\label{sec:exp}
		
		In this section, we will describe the details in the study to evaluate ChatGPT's capabilities spam detection tasks. This section covers the following main components: datasets, Prompt design, implementation of baseline models, and evaluation metrics. Both the English and Chinese datasets can be found in the github repository: \url{https://github.com/shijing001/spam_detection}.
		
		

		\subsection{Datasets}
		We used two baseline datasets: The open-sourced email spam detection dataset in English and a low-resourced Chinese dataset collected by the authors.
		\begin{itemize}
			\item \textbf{Email Spam Detection Dataset (ESD):}  It is a dataset contains 5572 rows, each row for each email.  Each line is composed of two columns: v1 contains the label (ham or spam) and v2 contains the raw text. This dataset is open-sourced in Kaggle \footnote{https://www.kaggle.com/datasets/shantanudhakadd/email-spam-detection-dataset-classification}.
			\item \textbf{Chinese Spam Dataset (CSD):} This data set is constructed by the authors, which are mainly collected from personal email messages and some spam messages are collected on the Internet. We label the corpus manually and cross-validate the labels to ensure the quality of the data.
		\end{itemize}

		\begin{table}[ht]
			\begin{center}
				\caption{\label{tab:data} Detail of Datasets }
				\begin{tabular}{|c|ccc|}
					\hline \bf Data & \bf Size & \bf Spam & \bf Ham \\ \hline
					ESD & 5572& 747& 4825\\
					CSD & 58& 44& 14\\
					\hline
				\end{tabular}
			\end{center}
		\end{table}
		
		The general information of the dataset is shown in Table \ref{tab:data}. This table displays the
		size of two datasets, and the number of spam and ham emails in these datasets.
		
		\subsection{In-context Learning}
		
		In-context learning was popularized in the original GPT-3 paper \citep{brown2020language} as a way to use language models to learn tasks given only a few examples in the instruction. During in-context learning, we give the language model a prompt that consists of a list of input-output pairs that demonstrate a task. At the end of the prompt, we append the email and ask the language model to make a prediction just by conditioning on the prompt and predicting the next tokens.
		To utilize ChatGPT via in-context learning,
		it is necessary to design an instruction, the so-called prompt.
		
		
		\subsubsection{Zero-shot Prompt}
		The output of ChatGPT is sometimes inconsistent, so we enhance the classification efficiency of the model by careful design of prompt \citep{cao2023study}. On top of the observation from real-world questions, we propose a zero-shot prompt template as follows:\\
		Forget all your previous instructions, pretend you are a junk mail detection expert who tries to identify whether a mail is spam mail. Answer \textbf{``spam"} if it is spam, \textbf{``ham"} if not. \textbf{And if you think it is too private or difficult to judge, you can exclude the impossible one and choose the other.}
		Now you can identify whether the mail is spam:\\
		Mail: [mail content]\\
		
		We design the prompt template in this way because ChatGPT sometimes has difficulty identifying email categories and returns responses like ``neutral" or ``ambiguous" under the zero-shot settings. Therefore, in order to achieve better binary classification, we utilize this prompt for the zero-shot experiments.
		
		\subsection{Baselines}
		We compared ChatGPT with popular mainstream spam detection models.
		\begin{itemize}
			\item \textbf{Naive Bayes:} It is a classification algorithm based on Bayes' theorem and feature independence assumption \citep{ma2020comparative}. It assumes that input features are independent of each other and uses prior probabilities of classes and conditional probabilities of features to calculate posterior probabilities for classification. Naive Bayes classifier is simple, computationally efficient, and widely used in text classification and spam filtering tasks. This method takes the TF-IDF features of texts as input.
			\item \textbf{Logistic Regression (LR):} It is a commonly used classification algorithm for solving binary classification problems. It builds a linear model and applies a logistic function (such as the sigmoid function) to perform classification predictions. Logistic regression can output the probability of a sample belonging to a certain class and make decisions based on a specified threshold. Its advantages include computational efficiency, simplicity, and interpretability. This method takes the TF-IDF features of texts as input.
			\item  \textbf{Support Vector Machine (SVM) \citep{khamis2020header}:} It is a popular classification algorithm that constructs an optimal hyperplane to separate different classes of samples. SVM maps data to a high-dimensional space and seeks a maximum-margin hyperplane to maximize the separation between classes. SVM performs well in handling high-dimensional data and non-linear classification problems, and different kernel functions can be used for different types of data. This method takes the TF-IDF features of texts as input.
			\item \textbf{Dense Neural Networks (DNN):} This method takes the word embeddings as input and use dense layers to extract hidden features and outputs the probability of being spam. We chose the Keras deep learning framework to implement the DNN model.
			\item\textbf{BERT:} It is a pre-trained language model based on the Transformer architecture. It is trained on large-scale unlabeled text to learn rich language representations. The BERT model can be used for text classification tasks by passing the input text representation to subsequent classification layers for prediction. The BERT model exhibits high performance in natural language processing tasks, particularly in handling contextual relevance and semantic understanding. This model takes the pre-trained embeddings as input.
		\end{itemize}

		\subsection{Metrics}
		For classification models, the commonly used evaluation metrics are precision, recall, accuracy and F1 score \citep{perez2012rough}.
		Following the literature, notations $TP$, $TN$, $FP$, $FN$ represent the number of true positive, true negative, false positive and false negative, respectively.
		The accuracy (Acc) refers to the proportion of correct predictions among the total number of predictions. It is calculated as follows:
		\[Acc = \frac{{TP+TN}}{{TN + TP+ FN+ FP}}.\]
		In the context of classification models, the recall (R) metric denotes the proportion of actual positive cases that are accurately identified by the model, which is defined as follows:
		\[R = \frac{{TP}}{{FN + TP}}.\]
		Conversely, the precision (P) metric represents the proportion of examples predicted as positive that are indeed truly positive, which is defined as follows,
		\[P = \frac{{TP}}{{TP + FP}}.\]
		Notably, the precision rate and recall rate often exhibit a trade-off phenomenon, where improvements in one metric may come at the expense of the other. This reciprocal relationship is particularly pronounced in large-scale datasets, necessitating a comprehensive consideration of both metrics. To strike a balance between these competing indices, the F1-score is commonly employed as a summary measure of model performance, which is defined as follows:
		\[F1 = \frac{{2\times~P\times~R}}{{P + R}}.\]
		
		Because the spam detection dataset is usually imbalanced, so we use macro-level precision, recall, F1 score to evaluate the performance of models. Additionally, accuracy is also an important metric.
		
		\subsection{Specific Settings}
		The experiment utilized the GPT3.5 API from OpenAI to conduct experiments. We provide a prompt as input, excluding any historical dialogue or context, and instructing the model to disregard any previous context in the prompt \citep{lopez2023can}. We removed all line breaks and kept the message length under 100 words, in case the whole length was too long (GPT has a limit of 2000 tokens total). We specifically required the ChatGPT to output ``ham" or ``spam",  without any explanation.
		
		For the ESD dataset, the train-test split ratio is 80-20, i.e., 80\% of data is used for training models and 20\% for testing. However, due to the small-size of CSD, the train-test split ratio is 50-50, i.e., half of data is used for training models, and the rest for testing.
		
		\section{Results and Analysis}\label{sec:res}
		
		\subsection{Main Findings}

		\begin{table*}[ht]
			\begin{center}
				\caption{\label{tab:res}A comparison of different methods for spam mail detection using  macro-level precision (Prec.), recall (Rec.), F1 score and accuracy (Acc.). The best and second-best results for each dataset and metrics are highlighted in bold and underlined, respectively.}
				\vspace{.2cm}
				\begin{tabular}{|l|cccc|cccc|}\hline
					\textbf{} & \multicolumn{4}{c|}{\textbf{ESD}} &\multicolumn{4}{c|}{\textbf{CSD}}\\
					&Prec. & Rec.&Acc.& F1& Prec. & Rec.& Acc.& F1\\\hline
					Naive Bayes&\underline{0.98}& 0.85& 0.96& 0.90 & 0.37& 0.50& 0.74& 0.42\\
					SVM& 0.97& 0.90&0.97& 0.93 &0.63& 0.55& 0.74& 0.56\\
					LR&  0.97& \underline{0.94}&\underline{0.98}& \underline{0.96}& 0.37& 0.52& 0.75& 0.43\\
					DNN&  \underline{0.98} & \underline{0.94}&\underline{0.98}& \underline{0.96}&0.39& 0.52& 0.76&  0.44\\
					BERT& \bf 0.99& \bf 0.98&\bf 0.99& \bf 0.98& 0.59& 0.54& 0.76&  0.56\\ \hline
					ChatGPT(0-shot)& 0.73& 0.86 &0.83& 	0.76&0.76& 0.81& {0.84}& {0.78}\\
					ChatGPT(1-shot)& 0.76& 0.87& 0.85& 	0.79& \underline{0.87}& \underline{0.74}&\underline{0.88}& \underline{0.78}\\
					ChatGPT(5-shot)& 0.77 & 0.88& 0.86& 	0.80& \bf 0.88&\bf 0.75&\bf{0.89}& \bf{0.79}\\
					\hline
				\end{tabular}
			\end{center}
		\end{table*}
		
		\begin{figure}[h!]
			\centering
			\includegraphics[width=.8\linewidth]{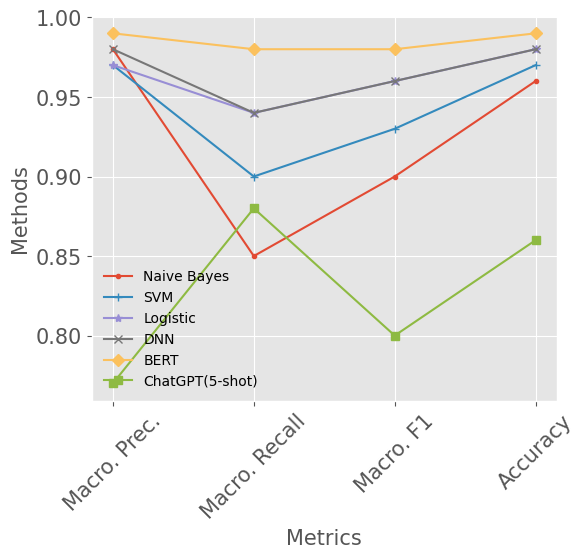}
			\caption{Comparison of performance of different methods on the English ESD dataset. The four evaluation metrics are
				macro-level precision, recall, F1 score and accuracy.}
			\label{fig:perf.en}
		\end{figure}
		
		This section exhibits the results from experiments and presents some analysis. Table \ref{tab:res} shows the performance of different models on both the English ESD and Chinese CSD datasets, in terms of macro-level precision, recall, F1 score and accuracy. From this table, BERT performs the best on the English ESD dataset, achieving the highest
		values on all four metrics, followed by DNN and LR. This result is pretty straightforward, as BERT is pre-trained and has the largest size of tunable parameters.
		In terms of macro F1 score, supervised learning methods (From naive Bayes to BERT) outperforms ChatGPT significantly, more than 10\%. Based on the ESD dataset, the spam detection ability of
		ChatGPT is significantly weaker than supervised learning models, especially BERT. Fig. \ref{fig:perf.en} illustrates the same information. This figure shows the performance of six methods on ESD in terms of four metrics (macro-level precision, recall, F1 score and accuracy). The ChatGPT with five-shot instances in the prompt is almost always below other lines.
		
		\begin{figure}[h!]
			\centering
			\includegraphics[width=.8\linewidth]{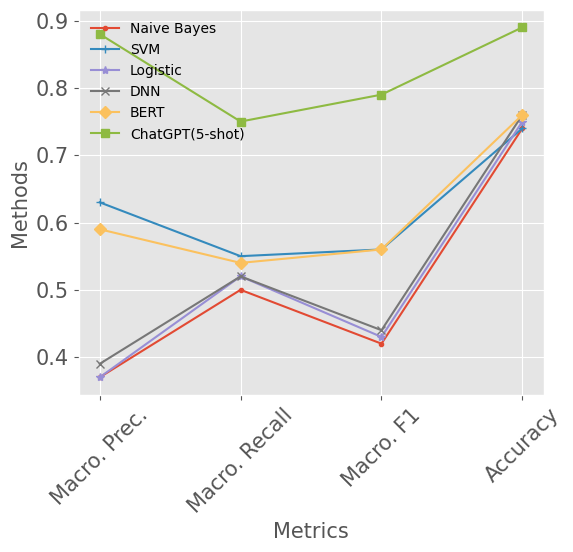}
			\caption{Comparison of performance of different methods on the low-resourced Chinese CSD dataset. The four evaluation metrics are
				macro-level precision, recall, F1 score and accuracy.}
			\label{fig:perf.cn}
		\end{figure}
		
		However, on the low-resourced Chinese CSD dataset, the story is quite different. The total sample size is 58, and 29 training examples and 29 testing examples. Due to the low-resourced training data, supervised learning methods perform poorly, BERT achieving a F1 score (0.56) and accuracy (0.76).
		The low-resourced situation does not affect the in-context learning of ChatGPT, and ChatGPT with five-shot instances performs the best on all four metrics. Figure \ref{fig:perf.cn} exhibits the same point, as ChatGPT lies above other models.
		
		By comparing the performance of ChatGPT on ESD and CSD datasets, we find that it performs consistently on both the English and Chinese datasets, ranging from 0.84 to 0.89 for accuracy and from 0.76 to 0.80 for F1 score.

		\begin{figure}[h!]
			\centering
			\includegraphics[width=.8\linewidth]{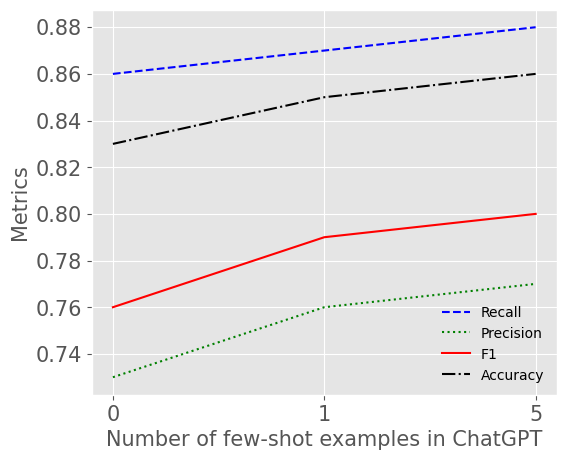}
			\caption{The performance of ChatGPT versus number of instances in prompts on the English ESD dataset}
			\label{fig:few.shot.en}
		\end{figure}

		\subsection{Effect of Instance Number in the Prompt}
		
		The performance of in-context learning
		depends on the number of instances/demostrations in the prompt.
		We designed the prompt template of few-shot as follows:\\
		Forget all your previous instructions, pretend you are a junk mail detection expert who tries to identify whether a mail is spam mail. Answer \textbf{``spam"} if it is spam, \textbf{``ham"} if not. And if you think it is too private or difficult to judge, you can exclude the impossible one and choose the other.\\
		\textbf{Here is a few examples for you:}\\
		$\bullet$  `Please don't text me anymore. I have nothing else to say.' is \textbf{``ham"};\\
		$\bullet$  `The New Jersey Devils and the Detroit Red Wings play Ice Hockey. Correct or Incorrect? End? Reply END SPTV' is \textbf{``spam"} .\\
		Now you can identify whether the mail is spam:\\
		Mail: [mail title]. [mail content]\\

		\begin{figure}[h!]
			\centering
			\includegraphics[width=.8\linewidth]{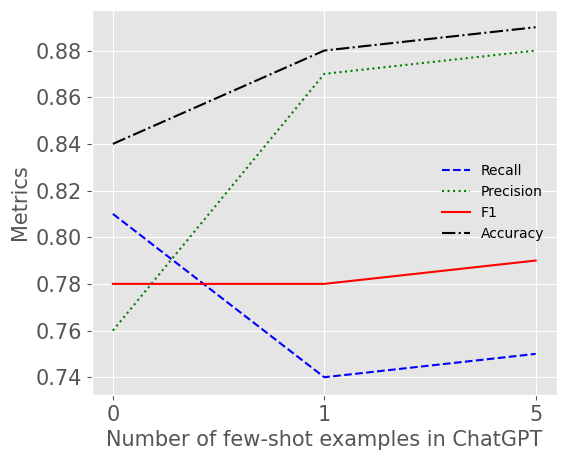}
			\caption{The performance of ChatGPT versus number of instances in prompts on the Chinese CSD dataset}
			\label{fig:few.shot.cn}
		\end{figure}
		
		Fig. \ref{fig:few.shot.en} and Fig. \ref{fig:few.shot.cn} present the effect of
		instance number on the performance of ChatGPT.
		From these two figures, generally the performance of ChatGPT becomes better in terms of all four metrics, as we increase the number of instance examples in the prompt.
		
		\begin{table}[h!]
			\begin{center}
				\caption{\label{tab:err}Examples of spam texts that incorrectly identified as ham. In the header, ``Quant." stands for ``quantitative information", and ``Pub info" for public information.}
				\begin{tabular}{|cc|cccc|}\hline
					Words&Length& Quant.& Number& Url& \makecell{Pub \\info}\\ \hline
					26&157&\checkmark &\checkmark& & \\
					29&154& &\checkmark& & \\
					26&155& &\checkmark&\checkmark& \\
					19&149& & &\checkmark & \\
					24&155& &\checkmark&\checkmark&\\
					16&92& & &&\checkmark\\
					22&159&\checkmark &\checkmark& & \\
					15&78& & & & \checkmark\\ \hline
				\end{tabular}
			\end{center}
		\end{table}
		
		\subsection{Error Analysis}

		In order to further elucidate the characteristics of misclassified emails, we conducted an in-depth examination of the textual data. Specifically, we undertook a comparative analysis of the text length and content of the misclassified emails. The results of this analysis are presented in Table \ref{tab:err}, which displays the details of eight spam emails that were incorrectly classified as ham. Notably, the text length of these misclassified emails tends to fall within the range of 150-160 words. However, no discernible pattern emerges in terms of word number distribution. A qualitative analysis of the text content reveals that six of the misclassified emails contain quantitative information, including prices and phone numbers. In contrast, the two remaining emails, which were shorter in length, contain references to public information, such as ``Divorce Barbie" and ``The New Jersey Devils", and employ a conversational tone reminiscent of small talk between acquaintances. As a result, these emails were mistakenly classified as ham.
		
		\subsection{Limitations}
		In this section, we will briefly summarize some of the limitations of the paper.
		
		\begin{itemize}
			\item \textbf{Evaluation Constraints:} Due to cost constraints, we only tested some of our baseline models on the mail classification task against zero-shot chat-gpt, and in fact, there may be other more representative machine learning and deep learning techniques that may provide additional perspectives for model evaluation. In addition, the GPT model in this paper uses the API of GPT-3.5, and with the development of large language models, large language models such as GPT-4 will also be put into formal use, so there are certain limitations in the performance evaluation of GPT in this paper.
			\item \textbf{Dataset Limitations:} In this paper, the self-constructed Chinese data set is selected as one of the main datasets, which is small in scale and needs to be expanded in the future. In addition, when studying the performance of different languages in the task of spam recognition, this paper only compares Chinese and English, which is representative to a certain extent, but there are still language limitations.
			\item \textbf{Input Length Limitation:} Because the GPT API has a limit on the length of tokens, and considering the uncertainty of the GPT's answers, this paper limits the input Prompt to 150 words (Email 100 words or less), so that the model can run normally. Therefore, the research of this paper mainly focuses on short texts with less than 100 words, and the research on long texts is still underexplored.
		\end{itemize}

		\section{Conclusion}\label{sec:concl}
		This paper explores the capabilities of ChatGPT in spam detection tasks and compares its performance with various baseline models such as SVM, LR, naive Bayes, and BERT. We evaluated the model on two datasets: the Email Spam Detection Dataset (ESD) and a low-resourced Chinese spam dataset.
		
		Our experimental results demonstrate that ChatGPT performs well in spam classification tasks, although its overall performance is significantly inferior to supervised models on the English ESD dataset. However, the spam detection ability of ChatGPT is pretty stable across different languages as its performance on the low-resourced Chinese CSD is similar to ESD.
		Another important observation is that
		ChatGPT has benefits on low-resourced datasets.
		
		\section*{Acknowledgments}
		
		The authors thank the reviewers for the
		valuable comments that helped to improve the paper. This work was supported
		by the Fundamental Research Funds for the Central Universities under Grant 2022114012,
		and the High-End Foreign Expert Recruitment Program of the Ministry of Science and Technology of the People's Republic of China, the construction of cutting-edge courses and academic research plans of International Finance and Big Data Management, under Grant G2023138001.
		

		\newpage
		\hbox to14cm{\hrulefill}

			\noindent\textsc{Shijing Si}\\
			School of Economics and Finance, Shanghai International Studies University\\
			Shanghai 201620, China\\
			E-mail Address: shijing.si@outlook.com\\
			
			\noindent\textsc{Yuwei Wu}\\
			School of Economics and Finance, Shanghai International Studies University\\
			Shanghai 201620, China\\
			E-mail Address: shakirawyw@gmail.com\\
			
			\noindent\textsc{Le Tang}\\
			Library, Shanghai International Studies University\\
			Shanghai 201620, China\\
			E-mail Address: 1067061406@qq.com\\

			\noindent\textsc{Yugui Zhang}\\
			School of Economics and Finance, Shanghai International Studies University\\
			Shanghai 201620, China\\
			E-mail Address: yuguizhang@126.com\\
			
			\noindent\textsc{Jedrek Wosik}\\
			Department of Medicine, Duke University\\
			Durham, North Carolina 27705, USA\\
            E-mail Address: jedrek.wosik@gmail.com\\

			\noindent\textsc{Qinliang Su}\\
1. School of Computer Science and Engineering, Sun Yat-sen
University; 2. Guangdong Key Laboratory of
Big Data Analysis and Processing\\
Guangzhou 510006, China\\
E-mail Address: suqliang@mail.sysu.edu.cn\\
	\end{document}